\documentclass[a4size,journal]{IEEEtran}
\usepackage{amsmath,amsfonts}
\usepackage{algorithmic}
\usepackage{algorithm}
\usepackage{array}
\usepackage[caption=false,font=normalsize,labelfont=sf,textfont=sf]{subfig}
\usepackage{textcomp}
\usepackage{stfloats}
\usepackage{url}
\usepackage{verbatim}
\usepackage{graphicx}
\usepackage{cite}
\usepackage{tcolorbox}
\usepackage{booktabs}
\usepackage{xcolor}
\usepackage{gensymb}

\usepackage{times}
\usepackage{epsfig}
\usepackage{amsmath}
\usepackage{amssymb}
\newcommand{\depth}{\blacklozenge}
\newcommand{\segm}{\mathsection}
\newcommand{\temporal}{\dagger}

\usepackage[pagebackref=true,breaklinks=true,letterpaper=true,colorlinks,bookmarks=false]{hyperref}
\begin{document}

\title{Addressing Limitations of State-Aware Imitation Learning for Autonomous Driving}

\author{\IEEEauthorblockN{Luca Cultrera\IEEEauthorrefmark{1}, Federico Becattini\IEEEauthorrefmark{2}, Lorenzo Seidenari\IEEEauthorrefmark{1}, Pietro Pala\IEEEauthorrefmark{1}, Alberto Del Bimbo\IEEEauthorrefmark{1}}\\~\IEEEmembership{\IEEEauthorrefmark{1}University of Florence, \IEEEauthorrefmark{2}University of Siena}
}



\maketitle

\begin{abstract}
Conditional Imitation learning is a common and effective approach to train autonomous driving agents. However, two issues limit the full potential of this approach: (i) the inertia problem, a special case of causal confusion where the agent mistakenly correlates low speed with no acceleration, and (ii) low correlation between offline and online performance due to the accumulation of small errors that brings the agent in a previously unseen state. Both issues are critical for state-aware models, yet informing the driving agent of its internal state as well as the state of the environment is of crucial importance.
In this paper we propose a multi-task learning agent based on a multi-stage vision transformer with state token propagation. We feed the state of the vehicle along with the representation of the environment as a special token of the transformer and propagate it throughout the network. This allows us to tackle the aforementioned issues from different angles: guiding the driving policy with learned stop/go information, performing data augmentation directly on the state of the vehicle and visually explaining the model's decisions.
We report a drastic decrease in inertia and a high correlation between offline and online metrics.
\end{abstract}


\section{Introduction}
\IEEEPARstart{A}{utonomous} driving is becoming a reality. To make this possible, several problems have to be solved, such as perception~\cite{pohlen2017full}, planning~\cite{claussmann2019review}, and forecasting~\cite{marchetti2020multiple}.
A recent trend that has obtained remarkable results is to directly train driving agents from raw observations with Imitation Learning (IL) \cite{codevilla2018end, codevilla2019exploring}, i.e. learning to mimic demonstrations from expert human drivers. In this way, the autonomous driving problem is tackled holistically, without having to rely on different heterogeneous modules.

Imitation learning, however, has some limitations. Since the driving capabilities are learned by behavioral cloning, IL models usually lack explicit causal understanding.
Rather than rules, relations between patterns are learned, thus making the agent vulnerable to spurious correlations in the data. This phenomenon is known in the literature as \textit{causal confusion} \cite{de2019causal}.
In particular, when training IL agents for automotive, there is evidence of a special case of causal confusion referred to as the \textit{inertia problem} \cite{codevilla2019exploring, greco2022imitation, samsami2021causal}. The inertia problem stems from a spurious correlation between low speed and no acceleration in the training data, making the driving agent likely to get stuck in a stationary state.
As a consequence, when a state-aware agent halts (e.g. at a traffic light or in a traffic jam), it may not move again when it should.
\textcolor{black}{For state-awareness here we refer to any source of information that can inform the agent about its halted state, such as a state variable, either explicitly modeled or implicitly inferred, that encodes velocity.}

A second issue that limits the applicability of IL is the gap between offline and online driving capabilities \cite{le2022survey, codevilla2018offline}. Codevilla \textit{et al.} \cite{codevilla2018offline} showed that there is a low correlation between offline evaluation metrics (e.g. frame-wise Mean Squared Error in steer angle prediction) and the success rate in online driving benchmarks.
In online driving, the output of the model influences future inputs, violating the i.i.d. assumption made by the learning framework \cite{ross2010efficient}. Accumulation of small errors thus brings the vehicle into new states, never observed at training time \cite{ross2011reduction}.
\textcolor{black}{Similarly to the inertia problem, this issue manifests itself the most in state-aware models: the more variables are observed by the model, such as ego-velocity or previous driving commands, the sparser the coverage of the training data gets, making it more likely to end up in under-represented configurations at driving time.}

To summarize, IL agents suffer from ill-distributed training data that presents spurious correlations and domain shift compared to the test set.
These issues make it particularly hard to train state-aware agents: using multiple input sources increases the chances of discovering unwanted correlations in the data or of observing under-represented inputs at inference time, for which the agent does not know how to act confidently \cite{ross2010efficient, ross2011reduction, schroecker2017state}. 
In this paper, we address these difficulties in training state-aware IL models.

In literature, some attempts to identify and solve these issues have been done.
The inertia problem has been addressed by regularizing training through vehicle speed prediction \cite{codevilla2019exploring}, whereas \cite{codevilla2018offline} demonstrated the usefulness of two data augmentation approaches to improve offline and online driving capability correlation: first, augmenting the training set with lateral cameras, thus simulating a vehicle with an unusual trajectory, and second, perturbing the driving policy to record samples where the vehicle recovers from anomalous states.
In this paper, we build on these ideas without collecting additional data. We propose an IL agent that propagates the state of the vehicle through the model and uses it as the core of a multi-task architecture.
On the one hand, this allows us to explicitly train the model to avoid issues such as the inertia problem. 
On the other hand, this allows us to perform data augmentation on all the observed data, reducing the distribution shift between training samples and what the agent may see at driving time.


Our IL agent is designed as a hierarchical transformer model with state token propagation. The vehicle's state is encoded in a special token of a vision transformer \cite{dosovitskiy2020image} and is enriched with new information at each stage of the architecture. At first, we predict whether the vehicle must stop or go, directly tackling \textit{inertia}. This information is passed to the next stage which predicts the driving commands (namely steer, throttle, and brake). Finally, the model leverages a differentiable Command Coherency Module (CCM), encouraging the model to correctly bring the vehicle to the desired future state by generating non-conflicting controls. \textcolor{black}{Such command is used only at training time and acts as a regularizer.}
Since our architecture is based on a transformer encoder \cite{vaswani2017attention}, it heavily relies on attention. We leverage such attention to gain insights about what the model is focusing on to make its decisions (e.g., the vehicle's state or visual patterns), following the recent trend of designing explainable driving models \cite{omeiza2021explanations, cultrera2020explaining, zablocki2021explainability}.

Interestingly, the ability to explain the model's decisions provides us with a better understanding of the inertia problem. Inertia makes an IL model halt and stay still whenever the speed of the vehicle is close to zero. However, it is hard to discriminate this phenomenon from other kinds of failures that make the vehicle stop indefinitely. For instance, if part of the environment is mistakenly interpreted as a crossing pedestrian or a red traffic light, the vehicle will wait indefinitely for the state of the surrounding environment to change.
Whenever this happens, a different solution must be sought in order to enforce the visual backbone of the model, rather than its causal inference capabilities.
By combining the model's attention with a retrieval-based explainability method, we are able to highlight these differences and isolate instances of inertia from backbone failures.

The main contributions of our paper are the following:
\begin{itemize}
\item We propose a state-aware conditional imitation learning model for autonomous driving. The model is multi-stage and exploits state propagation through different transformer layers breaking down the generation of driving commands into coarse to fine tasks.
\item We specifically address issues in state-aware imitation learning such as the inertia problem and the offline/online performance gap. Inertia is drastically limited by state token propagation and multi-stage learning, whereas the correlation between online success rate and offline metrics is enforced via data augmentation on the vehicle's state. 
\item We propose a combination of the transformer's self attention with an ex-post semantic explainability method that we use for inspecting model failures. This points out interesting "hallucinations" of the visual backbone that cause behaviors mistakenly confused with inertia.
\end{itemize}

\section{Related Works}

Imitation Learning (IL) is based on the idea that, to learn a complex task, a model can observe the demonstrations of an expert performing it \cite{argall2009survey, attia2018global}.
This paradigm has been successfully applied to autonomous driving. One of the first approaches based on IL predicted steering commands for lane following and obstacle-avoiding tasks \cite{bojarski2016end}.
The task soon evolved into the so-called Conditional Imitation Learning (CIL), in which predictions are conditioned on high-level commands such as \textit{turn} or \textit{go straight}.
Several works followed this approach~\cite{codevilla2018end, sauer2018conditional, xiao2020multimodal, cultrera2020explaining, haris2022navigating}, also combining it with reinforcement learning~\cite{liang2018cirl, toromanoff2020end, zhang2021end}.

To obtain better driving capabilities, several sensors and additional synthetic data are often used~\cite{codevilla2018end, lee2018context, berlincioni2021multiple, chen2020learning, yang2018end}. Large use of environmental information is done by prior work in the form of semantic segmentations~\cite{li2018ret, toromanoff2020end, chekroun2021gri, chen2022learning}, top-view maps \cite{chen2020learning}, or both \cite{chen2021learning, chitta2021neat, hu2022st, chitta2022transfuser, zhang2021end}.
Similarly, other methods leverage depth information~\cite{xiao2020multimodal, chitta2022transfuser}, LiDAR data~\cite{haris2022navigating, li2018ret, haris2022navigating, chitta2022transfuser} or cues such as traffic light states~\cite{toromanoff2020end, chekroun2021gri}, lane position~\cite{toromanoff2020end}, and intersection presence~\cite{toromanoff2020end, chekroun2021gri}.
Such data has been also used in the form of \textit{affordances}, low-dimensional representations of environmental attributes~\cite{sauer2018conditional, chen2015deep, toromanoff2020end}.
Differently from all the aforementioned methods, we rely on a purely RGB-based approach. Whereas these methods have access to environmental data, either as inputs or as additional sources of supervision, we assume to have access only to the RGB stream and the state of the vehicle (i.e., current speed, steer, acceleration, and brake), which is a direct consequence of the driving policy. A similar assumption is done in recent works such as~\cite{codevilla2019exploring, cultrera2023explaining, xiao2023scaling}.

Training a state-aware imitation learning agent hides some challenges~\cite{schroecker2017state}. Despite its simplicity and effectiveness, it breaks the i.i.d assumption made by any statistical supervised learning framework since current decisions influence future inputs~\cite{ross2010efficient, ross2011reduction}.
The main difficulty that needs to be addressed is trying to keep the model in a state close to what has been observed at training time \cite{schroecker2017state}. When this does not happen, online errors tend to accumulate over time, generating less accurate behaviors~\cite{pomerleau1988alvinn, ross2010efficient}.
The effect is to have online capabilities that do not correlate with offline error metrics measured on a validation set~\cite{codevilla2018offline}, which makes the agent difficult to train. A solution to bridge this gap is to perform data augmentation. Codevilla \textit{et al.}~\cite{codevilla2018offline} showed that collecting data from three different cameras while adding noise to the driving policy helps in recovering from unexpected scenarios.
This however requires collecting hours of additional data.
Image-level data augmentations such as changes in contrast,
brightness and tone also have beneficial effects, especially for generalizing to similar scenarios with different conditions (e.g. weather)~\cite{codevilla2018offline, codevilla2019exploring}. Nonetheless, augmenting the pixel space has a limited effect on state-aware models, where predicted quantities are provided as input.
Differently from prior work, we perform augmentation on the vehicle's state, injecting it into the model as a special token of a transformer~\cite{vaswani2017attention}. Augmenting the state leads to a better coverage of the state space during training.

The presence of the state token allows us to address another well-known issue with imitation learning in automotive: the inertia problem~\cite{codevilla2019exploring, greco2022imitation}. This has been addressed in literature by predicting the current speed of the vehicle~\cite{codevilla2019exploring} or via causal imitative learning~\cite{samsami2021causal}, also based on speed prediction. A memory-based approach for retrieving previously observed scenarios has also been exploited recently~\cite{chuang2022resolving}.
The common speed-prediction solution proposed in~\cite{codevilla2019exploring} suffers from a high collision rate, likely due to overcompensation of inertia.
Instead of making the network predict its current velocity, we leverage a multi-stage architecture, where a stop/go loss based on the actual causes for stopping (presence of pedestrians, traffic lights, other vehicles) conditions the command generation. In this way, we inform the model about external elements that should be taken into account while driving. We find this solution to almost eradicate inertia entirely.
\section{Overview}
\begin{figure*}[th]
  \centering
    \includegraphics[width = 0.85 \linewidth]{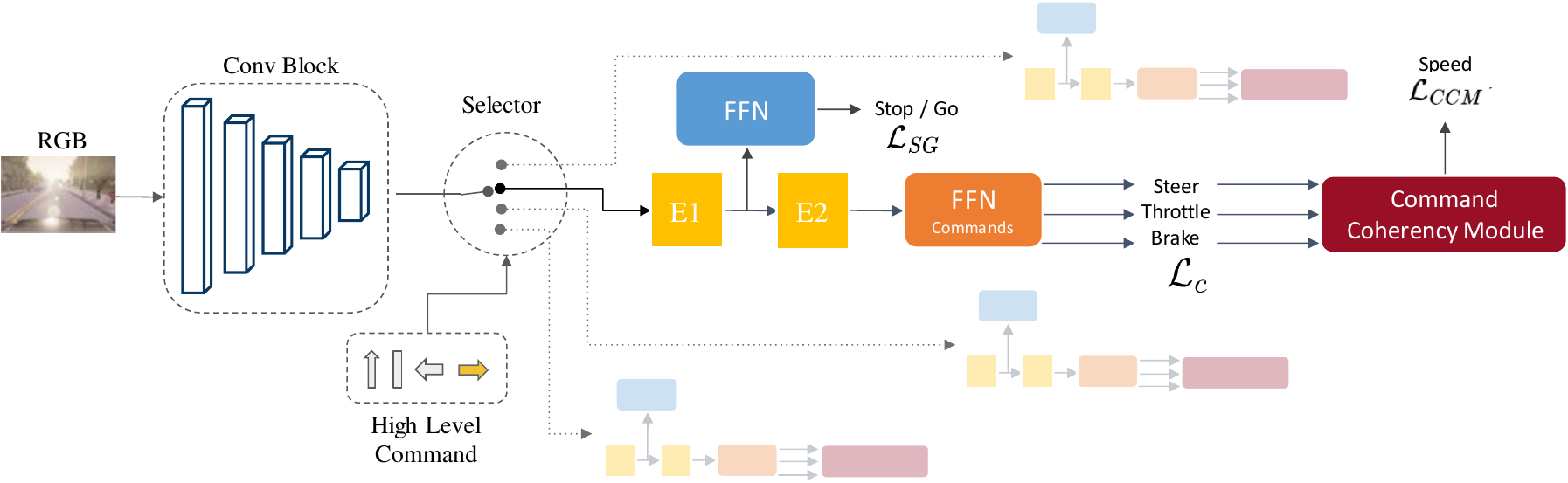}
  \caption{A convolutional backbone extracts a feature map, which is fed to a multi-stage transformer architecture. The first stage (E1) takes the feature and a state token, which is propagated across the network. The output of E1 corresponding to the state token is decoded into a stop/go prediction with a Feed Forward Network (FFN). The second stage (E2) uses the propagated state to predict driving commands. Finally, the Command Coherency Module is used as a loss regularizer to ensure consistency between driving commands.}
  \label{fig:architecture}
\end{figure*}

\label{sec:model}
Imitation Learning (IL) trains an agent by observing a set of expert demonstrations to learn a policy \cite{attia2018global}.
In the simplest scenario, IL is a direct mapping from observations to actions \cite{argall2009survey}.
In automotive, the expert is a driver, the policy is \textit{"safe driving"} and the demonstrations are a set of \textit{(frame, driving-controls)} pairs.
In this paper, we address Conditional Imitation Learning (CIL), a declination of imitation learning where the policy must reflect a given high-level command, such as \textit{turn right} or \textit{follow lane}.
As in prior work (e.g.~\cite{codevilla2018end, sauer2018conditional, cultrera2020explaining}), we divide our architecture into multiple branches, with separate heads learning command-specific policies. However, differently from prior work, we structure our model as a hierarchy of stages, each of which is dedicated to addressing different aspects of driving, as depicted in Fig. \ref{fig:architecture}.

The proposed model is state-aware, in the sense that it takes as input the speed and the steer, acceleration and brake values predicted at the previous timestep. In principle, informing the model of the current state of the vehicle could ensure temporal smoothness and coherency in the driving policy (i.e., the predicted driving controls).
In practice, this makes the model vulnerable to spurious correlations in the data, bringing out the \textit{inertia problem}.
To address this issue, we propose a multi-stage transformer model with state token propagation.
We feed the vehicle state to the model as a special token of a vision transformer (ViT) \cite{dosovitskiy2020image}. Operatively speaking, the state token fulfills the same role as the \textit{CLS} token in standard ViTs.
However, by enclosing vehicle measurements we can inject information into the model and let it correlate to relevant spatial features via self-attention. After each layer, the state token is enriched with spatial information and is decoded into coarse-to-fine driving commands, depending on the stage. The coarser of such commands is a decision on whether the vehicle should stop or go, thus explicitly addressing inertia.
Injecting the state token into the model has the additional benefit of enabling data augmentation on the state values itself, addressing what is arguably the biggest limitation of imitation learning, i.e., the inability to perform well in previously unseen states \cite{le2022survey} \textcolor{black}{that is also responsible for the gap of accuracy between offline and online driving}.
We also introduce a regularizer that ensures coherency in the generated driving commands. This is different from similar solutions adopted in prior works, where speed is predicted to reduce inertia \cite{codevilla2019exploring}, but here we use it to reduce online-offline evaluation gap.





\section{Model}



\subsection{State Token Propagation}
\label{sec:state_token}

Our model exploits a multi-stage transformer encoder architecture. The hierarchy of layers reflects a coarse-to-fine learning where each stage generates a different output. The rationale is that the $i-th$ stage can inform stage $i+1$ by taking the output of the encoder corresponding to the state token and propagating it as the new state token.
To enrich the token with increasingly complex semantics, at each stage we decode it into a different output with a Feed Forward Network (FFN), specific for separate tasks.

We define our multi-task hierarchy as follows.
The first stage predicts whether the agent should halt the car or keep it going. This is specifically thought to address the inertia problem. This stage does not produce any driving control and is expected to focus on traffic lights and other agents.
The second stage generates the actual driving commands: throttle, brake, and steer.
This second stage of the model should instead learn and understand road topology and ego-motion patterns. Thanks to the propagated state token, the generation of the driving commands is conditioned on the stop/go decision of the previous stage.
The third and final stage is the command coherency module that acts as a regularizer, thus we use it only at training time.
The initial state token is the embedding of steer, throttle, brake and speed at time $t-1$.

To cope with the non-uniform distribution of vehicle states in the train set (see Sec. \ref{sec:noiseinj}), we introduce a data augmentation strategy based on noise injection to perturb the state token.
We inject a zero mean Gaussian noise with $\sigma=0.1$ for driving controls, since they are all in $[0,1]$. For the speed, that takes values in $[0,10]$, we use $\sigma=1$ instead. 

\subsection{Pixel-State Attention}
\label{sec:visual_attention}
Every stage of the model performs token-to-token attention, thanks to the transformer's self-attention.
The advantages are twofold: on the one hand, prior work has shown that explicitly modeling attention improves driving capabilities \cite{cultrera2020explaining, cultrera2023explaining}; on the other hand, it provides a built-in interpretability mechanism that can be used to visually explain decisions. 

In our model, the attention involves not only visual patches as in \cite{cultrera2020explaining, cultrera2023explaining} but also the state of the vehicle.
First, the output of the convolutional backbone, i.e. a feature map $f$ of size $H_f \times W_f \times C$, is flattened into $N = H_f~\cdot~W_f$ separate $C$-dimensional tokens, corresponding to $1 \times 1$ spatial patches in the feature map.
Patches are then linearly projected into a $D$-dimensional space to adapt them to the input size of the transformer.
The four scalar quantities that compose the state of the vehicle (\textit{speed}, \textit{steer}, \textit{acceleration} and \textit{brake}) are lifted to a dimension of $D/4$ and concatenated into the $D$-dimensional state token, which we refer to as $x_{state}$. As in \cite{dosovitskiy2020image}, a learnable positional embedding $E_{pos}$ is added to all the $N+1$ tokens.
To summarize, the set of $N+1$ tokens fed to the encoder is composed as follows:
\begin{equation}
    z = [x_{state}; f^1 P_v ; \ldots ; f^N P_v] + E_{pos}
    \label{eq:eq1}
\end{equation}
where $P_v \in \mathbb{R}^{C \times D}$ is the feature projection matrix, $E_{pos} \in \mathbb{R}^{(N+1) \times D}$ and $x_{state} \in  \mathbb{R}^{1 \times D}$.

The self-attention carried out in every layer of the transformer is thus a pixel-state attention, where every pixel of the feature map can attend to each other plus the state token. This allows us to inspect at each stage which information is privileged by the model: when the state token carries relevant information from the previous stage (e.g., if the vehicle must stop), the model will give it high importance; vice-versa, if the image carries meaningful cues (e.g., an intersection) the model will focus on the interested pixels.

\subsection{Command Coherency Module}
\label{sec:CCM}

A possible cause for low correlation between off-line error and on-line driving performance~\cite{codevilla2018offline} can be found in throttle, brake and steer being predicted independently. What is missing is the optimization of a common goal that brings the vehicle from one initial state to a desired one, considering all three quantities. Furthermore, individual biases may interfere with the quality of the overall policy.

To generate the appropriate driving behavior, the predicted commands must be compatible with each other.
To this end, we introduce the Command Coherency Module~(CCM).
The CCM takes as input steer, throttle, brake and speed at time $t$ and predicts the future speed at time $t+1$.
We first train the command coherency module on training measurements to learn how such quantities affect the speed of the vehicle. Once the module is trained, we freeze it and use it as a regularizer while training the driving agent.
To implement the CCM, we use a lightweight multi-layer perceptron with three layers and ELU activations.

Our CCM shares some traits with the speed prediction module of \cite{codevilla2019exploring}. Here, the authors feed a frame-based estimate of the speed to the model. Instead of feeding the predicted speed as an additional input, we optimize it to regularize the outputs and conciliate the driving commands.

\subsection{Architecture and Training}
\label{sec:traindetails}
The proposed model is composed of a shared convolutional backbone plus four parallel branches, one for each high-level command.
The shared backbone consists of 5 convolutional layers with ELU activations. The first three layers have respectively 24, 36, and 48 $5 \times 5$ kernels with stride 2, followed by two other layers with 64 $3 \times 3$ filters with stride 1.
Input images are resized to a $200 \times 88$ px, yielding a $4 \times 18 \times 64$ feature map. After flattening we obtain $N=72$ visual tokens.
Each branch is a multi-stage transformer encoder with input size $D=64$. We use $3$ heads with a depth of $4$ for each encoder stage.

The first stage of the transformer takes the state token $x_{state}$ along with the $N$ visual tokens. The stage outputs $N+1$ transformed tokens, among which the enriched state token is used to predict whether the vehicle should stop or go. To optimize the stop/go prediction we use an L1 loss:
\begin{equation}
    \mathcal{L}_{SG} = \frac{\vert S_{TL} - \bar{S}_{TL} \vert +  \vert  S_{P} - \bar{S}_{P} \vert + \vert  S_o - \bar{S}_{o} \vert}{3} 
    \label{eq:eq4}
\end{equation}
where $S_{TL}$, $S_{P}$, $S_o$ represent intention signals in $[0, 1]$ \cite{codevilla2019exploring}, respectively for traffic light stop, pedestrian stop, and stop due to other vehicles.

The second stage is in charge of generating driving commands. Similarly to the first stage, the propagated state token taken from the output of the stage is fed to a feed-forward regressor to predict steer, throttle and brake.
We use an L1 loss for driving command prediction:
\begin{equation}
    \mathcal{L}_{c} = \vert \alpha (s - \bar{s}) \vert + \vert \beta (t - \bar{t}) \vert + \vert \gamma (b - \bar{b}) \vert 
    \label{eq:eq2}
\end{equation}
where $s \in [0,1]$, $t \in [0,1]$ and $b \in [0,1]$ are respectively the predicted steer, throttle and brake values, $\bar{s}$, $\bar{t}$ and $\bar{b}$ the corresponding ground truth values and $\alpha$, $\beta$, and $\gamma$ are weights with values  $0.5, 0.45$, and $0.05$, as in \cite{xiao2020multimodal}.
For the Command Coherency Module, we also use an L1 loss.
\textcolor{black}{The CCM loss $\mathcal{L}_{CCM}$, and the stop/go loss, denoted as $\mathcal{L}_{SG}$, contribute to the total loss according to: $\mathcal{L}_{Total} = \lambda \mathcal{L}_{c} + \kappa \mathcal{L}_{CCM} + \tau \mathcal{L}_{SG}$  where $\lambda=0.8$,  $\kappa=0.1$ and $\tau=0.1$.}
We train our model end to end with the Adam optimizer for $100$ epochs with a batch size of $64$ and a learning rate of $0.0001$.
\section{Model Explainability}
\label{sec:explainability}

The self-attention of the transformer stages in our model allows us to inspect the behavior of the model, thus providing explanations for the predictions. We refer to this as \textit{built-in explainability}. Since we have dedicated each stage of the model to different tasks, we can leverage such information to gain insights about what is important for different aspects of the learned policy.
We combine the built-in explainability with \textit{ex-post explainability}, i.e. an approach specifically designed to provide an additional interpretation of the model's behavior at inference time.

\subsection{Built-in Explainability} In both stages of the transformer model, we can obtain visual explanations in the form of attention maps.
The maps are obtained by considering the attention between the state token and the image patches.
The first stage provides information on what the model looks at for stop/go prediction, whereas the second identifies relevant image regions for a correct navigation.

\subsection{Ex-Post Semantic Explainability}
Built-in explainability only explains which regions are taken into account. However, it does not provide information about how these regions are interpreted by the model.
We propose an ex-post semantic explainability that combines visual attention with k-NN search of image features.

We gather offline a set of $m$ feature maps from the training set and collect the $D$-dimensional descriptors of each spatial location. In this way, we obtain a total of $M=m*N$ feature vectors, $N$ being the number of image spatial patches. We denote the i-th feature in the set as $y_i$.
At inference time, we extract the feature map $f$ of the input image and, for any spatial location of interest (e.g., the most attended ones by built-in attention), we perform a k-NN search with FAISS \cite{johnson2019billion} using the $L2$ distance:
\begin{equation}
    L = \textit{k-argmin}_{i=1:M} \Vert f_p -y_{i} \Vert_2
\end{equation}
where $f_p$ is the p-th feature vector of the input image ($p \in \{1, \ldots, N\}$).

For each k-NN we reproject the feature back onto the original image and take the semantic segmentation of the corresponding region\footnote{Ground truth segmentations are available in the \textit{NoCrash} dataset~\cite{codevilla2019exploring}}.
This allows us to inspect what the model is hallucinating by finding the dominant semantic category in the neighbors and allows us to interpret failures.


\begin{table}[t]
\centering
\caption{Success rate on \textit{Corl2017}. Superscripts identify additional data sources: $\depth$ (depth), $\segm$ (semantic segmentation), $\temporal$ (temporal modeling), $\star$ (different training data).
}
\resizebox{0.99\linewidth}{!}{
\begin{tabular}{lcccccc}
\hline
Model& \begin{tabular}{@{}c@{}}Training \\ Conditions\end{tabular} & & \begin{tabular}{@{}c@{}}New \\ Weather\end{tabular} & \begin{tabular}{@{}c@{}}New \\ Town\end{tabular} & \begin{tabular}{@{}c@{}}New Town\\ and Weather\end{tabular} & Overall\\\hline
MP$^{\segm}$  \cite{dosovitskiy2017carla} &  84.2 &  & 94.5 &51.2 &47.7  &69.4 \\
MT$^{\segm \depth}$ \cite{li2018ret} &  86.7 & & 89.0 &76.5&79.5 &83.0\\
CAL$^{\temporal}$ \cite{sauer2018conditional} & 93.0 &  & 92.0&77.2 &74.5 &84.2\\
EF$^{\depth}$ \cite{xiao2020multimodal} & 94.7 & &  92.0 & 88.5&91.0 &91.5 \\ 
CEF$^{\depth}$ \cite{haris2022navigating} & 94.5 & &  87.2 & 87.2&91.0 &90.0\\
MTL$^{\segm \depth}$ \cite{ishihara2021multi} & 99.7 & & 98.2 & 95.2&96.2 & 97.3\\
CILRS$^{\star}$ \cite{codevilla2019exploring} & 93.7 & &  $96.0$ & 78.7&92.5 & 90.2\\
LBC$^{\star}$ \cite{chen2020learning}& 100 & & 99.0 & 99.7 & 100 &  99.6\\
\hline
RL \cite{dosovitskiy2017carla} & 86.0 & & 26.5& 22.7 &24.5 & 40.0\\
CIL \cite{codevilla2018end} & 88.2& & 88.5 &58.5&53.5 & 72.2\\
EF-RGB \cite{xiao2020multimodal} &90.5 & & 75.5 &67.7&65.0 & 74.7\\
CIRL \cite{liang2018cirl} &92.5& & 90.0&66.2 &\textbf{80.0} & 82.2\\
LVA \cite{cultrera2020explaining}&  93.7 &  &  96.0&67.7&77.7 & 83.8\\
TRP \cite{cultrera2023explaining}& 96.2 & & 97.0&  72.5 & 73.5 & 84.8\\
Ours & \textbf{96.2} & & \textbf{97.2} &\textbf{78.0}&78.0 & \textbf{87.4}\\ \hline
\end{tabular}
}
\label{tab:Transf_exp}
\end{table}

\begin{table}[t]

\caption{Success rate on \textit{NoCrash}. Methods above the line have access to privileged information not used by other RGB-based methods: top-view maps (LBC, WOR), semantic segmentations (LBC, WOR) and additional data (FASNet, MoDE, CADRE, GRIAD).}
\centering
\resizebox{0.99\columnwidth}{!}{
\begin{tabular}{lccclccc}
\toprule
&  \multicolumn{3}{c}{Training conditions} &  & \multicolumn{3}{c}{New Weather}\\
\midrule

Model & Empty & Regular & Dense & & Empty & Regular & Dense \\
\midrule

LBC \cite{chen2020learning} & 89 & 87 & 75 & & 60 & 60 & 54 \\
FASNet \cite{kim2020multi} & 96 & 90 & 44 & & 98 & 80 & 38 \\
MoDE \cite{kim2022learning}  & 98 & 93 & 45 & & 98 & 84 & 46 \\
WOR \cite{chen2021learning} &  98 & 100 & 96 & & 90 & 90 & 84 \\
CADRE \cite{zhao2022cadre} & 95 & 92 & 82 & & 94 & 86 & 76 \\
GRIAD \cite{chekroun2021gri} & 98 & 98 & 93 & &  83 & 86 &  82 \\
\hline
CIL \cite{codevilla2018end} & 79 & 60 & 21 & & 83 & 55 & 13 \\
CAL \cite{sauer2018conditional} & 81 & 73 & 42 & & 85 & 68 & 33 \\
MT \cite{li2018ret} & 84 & 54 & 13 & & 58 & 40 & 7 \\
HDRM \cite{greco2022imitation} & 80 & 70 & 26 & & 92 & 76 & 44 \\
CILRS \cite{codevilla2019exploring} & 97 & 83 & 42 & &  \textbf{96} & 77 & 47  \\
Ours & \textbf{98} & \textbf{90} & \textbf{50} & &  90 & \textbf{88} & \textbf{52} \\
\bottomrule


\toprule

& \multicolumn{3}{c}{New Town} &  & \multicolumn{3}{c}{New Weather \& New Town}\\

\midrule

Model & Empty & Regular & Dense & & Empty & Regular & Dense \\
\midrule

LBC \cite{chen2020learning} & 86 & 79 & 53 & & 36 & 36 &  12 \\
FASNet \cite{kim2020multi} & 95 & 77 & 37 & & 92 & 66 & 32 \\
MoDE \cite{kim2022learning}  & 93 & 80 & 37 & & 94 & 68 & 34 \\
WOR \cite{chen2021learning} & 94 & 89 & 74 & & 78 & 82 & 66 \\
CADRE \cite{zhao2022cadre} & 92 & 78 & 61 & & 78 & 72 & 52 \\
GRIAD \cite{chekroun2021gri} & 94 & 93 & 77 & &  68 & 63 &  52 \\
\hline
CIL \cite{codevilla2018end} & 48 & 27 & 10 & & 24 & 13 & 2 \\
CAL \cite{sauer2018conditional} & 36 & 26 & 9 & & 25 & 14 & 10 \\
MT \cite{li2018ret} & 41 & 22 & 7 & & 57 & 32 & 14 \\
HDRM \cite{greco2022imitation} & 27 & 16 & 6 & & 8 & 4 & 2 \\
CILRS \cite{codevilla2019exploring} & 66 & 49 & 23 & &  \textbf{90} & 56 & 24  \\
Ours & \textbf{72} & \textbf{54} & \textbf{32} & &  84 & \textbf{58} & \textbf{28} \\

\bottomrule
\end{tabular}

}
\label{tab:tab_noCrash}
\end{table}

\section{Experimental results}
\label{sec:results}


\subsection{Dataset}
\label{sec:dataset}
For training and evaluating our model, we use the  \textit{Corl2017}~\cite{dosovitskiy2017carla} and NoCrash~\cite{codevilla2019exploring} datasets, both based on the Carla simulator~\cite{dosovitskiy2017carla}.
The \textit{Corl2017} dataset has expert demonstrations driving across the same town with a set of different weather conditions. Testing is performed by driving in different conditions: same town and weather as training; same town and new weather; new town; new town new weather. Testing also includes 4 tasks: go straight, one turn, navigation, navigation dynamic. The navigation tasks require driving from two distant waypoints and the dynamic scenario includes other vehicles and pedestrians.
\textit{NoCrash} has been designed to evaluate advanced driving skills such as stopping at traffic lights, avoiding collisions and driving in dense traffic environments.
The evaluation involves 25 episodes on three navigation tasks, spanning from an empty town scenario to a dense traffic one.
\textit{Corl2017} has 657.601 frames and \textit{NoCrash} instead 1.279.738 frames, divided into frontal and two lateral cameras ($-30\degree$,$+30\degree$).
For both datasets, the agent must comply with a given high-level command among \textit{go straight}, \textit{turn right}, \textit{turn left} and \textit{follow lane}.
As in~\cite{codevilla2019exploring} we train on a subsample of 10\% of the data, comprising 10 hours out of a total 100 hours of driving.
Both datasets provide meta-data including the current state of the autonomous vehicle and environmental information such as driving commands, high-level commands and position.

\subsection{Results}
\label{sec:results}
We report in Tab.~\ref{tab:Transf_exp} the results on the \textit{Corl2017} dataset. For a fair analysis, we compare our method directly against other RGB-based methods. We also report methods that leverage additional sources of supervision such as depth and semantic segmentation or additional data to train the model. The results show that our method obtains better or on-par results when compared to other RGB-based models. Per-task success rates are in the supplementary material.

Compared to \textit{Corl2017}, where traffic light violations and collisions are not considered, the \textit{NoCrash} benchmark is extremely more challenging since environmental cues must be taken into account. We report results in Tab.~\ref{tab:tab_noCrash}.
Our approach outperforms RGB methods, with the only exception of CILRS \cite{codevilla2019exploring}, which performs slightly better in some empty scenarios. In the more challenging scenarios with regular and dense traffic, our approach performs better than the competitors, highlighting the capacity of the model to interpret patterns relative to traffic lights and other agents.

In Tab. \ref{tab:TL_violations} we show the percentage of traffic light violations committed by our model. These results are computed on the task \textit{Empty} both for \textit{Training Conditions} and for \textit{New Weather \& New Town}.
As a baseline, we also report the results for a Single Stage model, i.e. a simplified version of our approach without the first stage. This model is state-aware as the full model, but does not exploit the stop/go loss which we designed to prevent inertia.
Interestingly, our model outperforms the single stage baseline by a large margin, showing the usefulness of the stop/go loss to correctly focus on traffic lights. 
At the same time, we significantly lower the traffic light violations compared to CILRS, despite it obtained a higher success rate in the empty tasks for \textit{New Weather \& New Town} (see Tab. \ref{tab:tab_noCrash}). We attribute this difference to two factors: (i) CILRS' strong ResNet vision backbone yields better generalization across weather conditions; (ii) higher capacity of our model to focus on traffic lights thanks to attention and stop/go loss.

\begin{figure}[t]
    \centering
    \setlength{\tabcolsep}{1pt}
    \begin{tabular}{ccc}

\includegraphics[width=.32\columnwidth ]{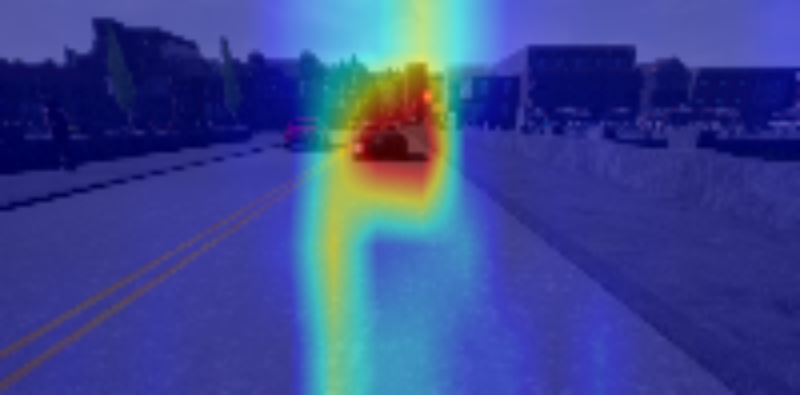} &
 \includegraphics[width=.32\columnwidth ]{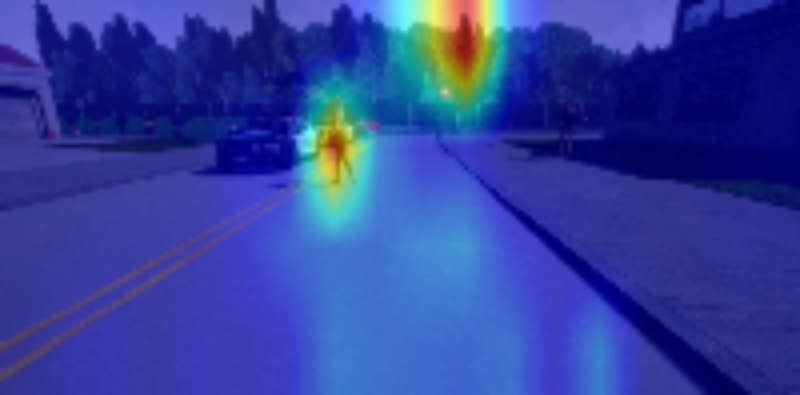} &
 \includegraphics[width=.32\columnwidth ]{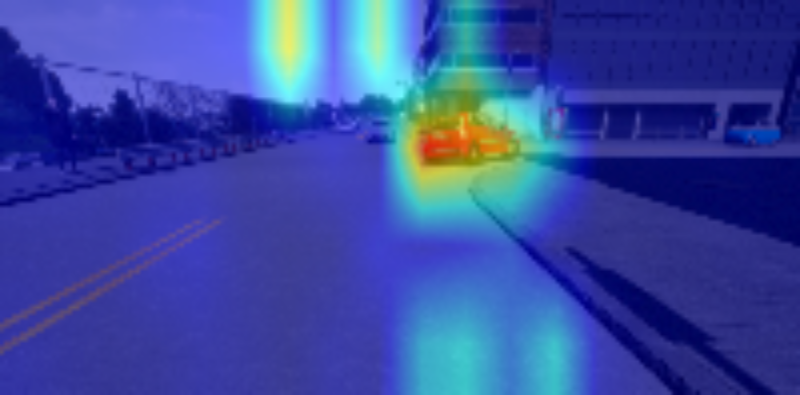} 
 \\
  \includegraphics[width=.32\columnwidth ]{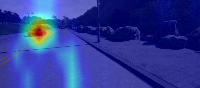} &
 \includegraphics[width=.32\columnwidth ]{Figures/Att/att_2.png} &
  \includegraphics[width=.32\columnwidth ]{Figures/Att/att_7.png} 
 \\
  \includegraphics[width=.32\columnwidth ]{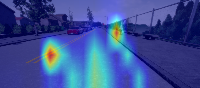} &
 \includegraphics[width=.32\columnwidth ]{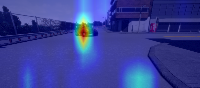} &
  \includegraphics[width=.32\columnwidth ]{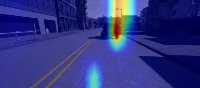} 
 \end{tabular}
  \caption{Examples of Stage 1 attention. This layer is dedicated to understanding whether the vehicle should halt its motion or not and therefore focuses on relevant cues in the scene such as vehicles, pedestrians and traffic lights.}
  \label{fig:suppAttStage1}
 \end{figure}

Attention plays an important role in identifying relevant cues. Since we employ transformer encoders in every stage of the model, we can visually inspect self-attention for every stage. We create heatmaps by reprojecting on the image the attention value relative to the state token against every visual token (Fig. \ref{fig:qualitative_attention}).
The heatmaps for the two stages reflect the tasks that are addressed at the corresponding levels: stop/go decision and driving command generation. The first stage focuses on small scene details such as traffic lights or pedestrians (additional qualitative examples for the first stage of our model are shown in Fig.~\ref{fig:suppAttStage1}), while the second stage attention is scattered and attends regions that are important for correct navigation such as intersections and roadsides.

\begin{table}[t]
\caption{Traffic light violations on \textit{NoCrash} benchmark (\textit{Empty}).}
\centering
\resizebox{0.85\linewidth}{!}{
\begin{tabular}{@{}c|ccc@{}}
\toprule
\multicolumn{3}{c}{Traffic Light Violations} \\ 
\midrule
Task & Train Conditions & New Town \& Weather  \\ 
\midrule 
CIL \cite{codevilla2018end}    & 83\%  &  82\%  \\
CILRS \cite{codevilla2019exploring}   & 27\%   & 64\%      \\
Single Stage   &  38\%   &  75\%  \\
Ours   &  \textbf{14}\%   &  \textbf{52\%}   \\
\bottomrule
\end{tabular}
}
  \label{tab:TL_violations}
\end{table}

\begin{table*}[t]
\centering
\caption{Ablation study switching off model components on \textit{NoCrash}. We remove: Noise Injection function (w/o NI), Command Coherency Module (w/o CCM), state token (w/o ST) and stop/go loss (w/o S/G loss). We also show results for the single stage baseline.}
\resizebox{0.95\textwidth}{!}{
\begin{tabular}{ccccccc|ccccccc}
\toprule
& \multicolumn{6}{c}{Training conditions} &  & \multicolumn{6}{c}{New Weather}\\
\midrule
Task & Single Stage  & w/o NI & w/o ST & w/o S/G loss & w/o CCM & Ours  &  &   Single Stage  & w/o NI & w/o ST & w/o S/G loss & w/o CCM & Ours\\
\midrule
Empty   & 73    & 94 & 82 & 80 & 84 & \textbf{98} &  & 72   & 90  & 62 & 84 & 80 & \textbf{90}\\
Regular & 60    & 88 & 50 & 56 & 60 & \textbf{90} &  & 56   & 72 & 46 & 46 & 52 & \textbf{88} \\
Dense   & 30    & 48 & 20 & 28 & 28 & \textbf{50}&  & 32  & 36  & 16 & 24 & 28 & \textbf{52}  \\
\midrule

& \multicolumn{6}{c}{New Town} &  & \multicolumn{6}{c}{New weather and new Town}\\
\midrule
Task & Single Stage  & w/o NI & w/o ST & w/o S/G loss & w/o CCM & Ours  &  &   Single Stage  & w/o NI & w/o ST & w/o S/G loss & w/o CCM & Ours\\
\midrule
Empty   & 60    & 70 & 60 & 58 & 64 & \textbf{72} &  & 62   & 72  & 66 & 64 & 70 & \textbf{84}\\
Regular & 36    & 48 & 24 & 28 & 32 & \textbf{54} &  & 32   & 40 & 16 & 24 & 36 & \textbf{58} \\
Dense   & 16    & 28 & 12 & 16 & 16 & \textbf{32}&  & 12  & 16  & 8 & 12 & 12 & \textbf{28}  \\
\bottomrule
\end{tabular}
}
\label{tab:ablations}
\end{table*}

\newcommand{\figw}{.32\linewidth}
\begin{figure}[t]
   \centering
   \begin{tabular}{@{}c@{~}c@{~}c@{}}
 RGB & Stage 1 & Stage 2 \\

\includegraphics[width=\figw, trim={15px 0px 15px 0px},clip ]{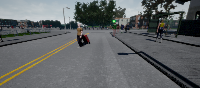} &
\includegraphics[width=\figw, trim={15px 0px 15px 0px},clip ]{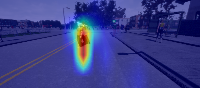} &
\includegraphics[width=\figw, trim={15px 0px 15px 0px},clip ]{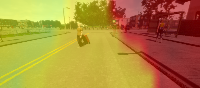} \\

\includegraphics[width=\figw, trim={15px 0px 15px 0px},clip  ]{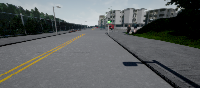} &
\includegraphics[width=\figw, trim={15px 0px 15px 0px},clip  ]{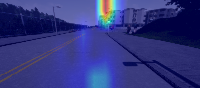} &
\includegraphics[width=\figw, trim={15px 0px 15px 0px},clip ]{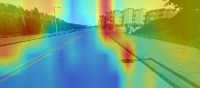} \\

\includegraphics[width=\figw, trim={15px 0px 15px 0px},clip ]{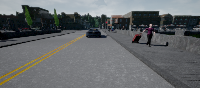} &
\includegraphics[width=\figw, trim={15px 0px 15px 0px},clip  ]{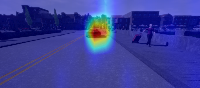} &
\includegraphics[width=\figw, trim={15px 0px 15px 0px},clip ]{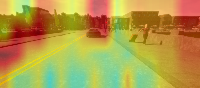}

\end{tabular}
\caption{Each row shows visual attention for the two stages of the model w.r.t an input image. The two stages reflect important cues for the stop/go and command generation losses respectively.}
\label{fig:qualitative_attention}
\end{figure}

\begin{figure}[t]
 \centering
\includegraphics[width=.95\linewidth,trim={0 18cm 0 0},clip]{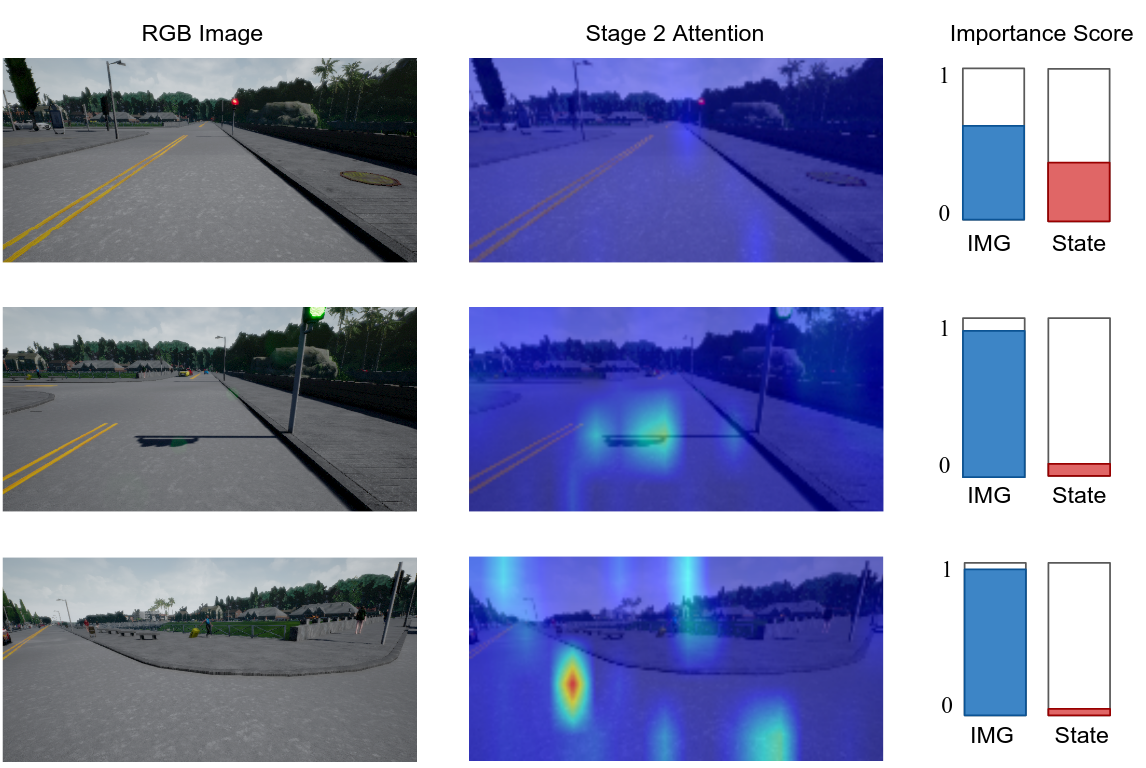}
\includegraphics[width=.95\linewidth,trim={0 9cm 0 10cm},clip]{Figures/SeqStateToken.png}
\includegraphics[width=.95\linewidth,trim={0 0 0 19cm},clip]{Figures/SeqStateToken.png}
\caption{Importance of state token vs image tokens. \textcolor{black}{The presence of a red traffic light is detected by the first transformer stage and encoded in the state token propagated to the second stage. In this case, the second transformed stage assigns a high attention value (red bar) to the state token}.
When restarting at the green light and turning, the image tokens (blue bar) gain importance.}
\label{fig:StateToken}
\end{figure}

\begin{table}[t]
\centering
\caption{Failure rate due to inertia problem in Town01 - New weather of the \textit{NoCrash} benchmark}
\resizebox{0.7\linewidth}{!}{
\begin{tabular}{@{}c|ccccc@{}}
\toprule
\multicolumn{4}{c}{Train conditions}  & \multicolumn{2}{c}{New weather}\\ 
\midrule
Task & Single Stage & Ours  & &Single Stage & Ours \\ 
\midrule 
Empty     & 17\%  &  \textbf{1\%} & & 40\% & \textbf{8}\% \\
Regular    & 9\%   &  \textbf{1\%}   & &20\%& \textbf{2\%}         \\
Dense   &  16\%   &  \textbf{6\%}  & & 22\%& \textbf{4\%} \\
\bottomrule
\end{tabular}
}
\label{tab:inertia}
\end{table}

\begin{table*}[t]
\caption{Success rate and number of failed episodes termination cause on NoCrash. We compare with the baseline and CILRS in all tasks and weather conditions in Town01. For success rate we underline the method with best performances (higher is better $\uparrow$), for failure cases we highlight (bold) the method with the lowest number of occurrences for that type of failure (lower is better $\downarrow$).}
\centering
\resizebox{0.99\textwidth}{!}{
\begin{tabular}{@{}lcccc|cccc|cccc|cccc|cccc|ccc@{}}
\toprule
\multicolumn{12}{c}{Train Conditions} & \multicolumn{1}{l}{} & \multicolumn{11}{c}{New Weather} \\ \midrule
\multicolumn{1}{c}{} & \multicolumn{3}{c}{Empty} & \multicolumn{1}{l}{} & \multicolumn{3}{c}{Regular} & \multicolumn{1}{l}{} & \multicolumn{3}{c}{Dense} & \multicolumn{1}{l}{} & \multicolumn{3}{c}{Empty} &\multicolumn{1}{l}{}  & \multicolumn{3}{c}{Regular} & \multicolumn{1}{l}{} & \multicolumn{3}{c}{Dense} \\\midrule
 & CILRS\cite{codevilla2019exploring} & Single Stage & Ours &  & CILRS\cite{codevilla2019exploring} & Single Stage & Ours &  & CILRS\cite{codevilla2019exploring} & Single Stage & Ours & & CILRS\cite{codevilla2019exploring} & Single Stage & Ours &  & CILRS\cite{codevilla2019exploring} & Single Stage & Ours &  & CILRS\cite{codevilla2019exploring} & Single Stage & Ours \\ \hline
Success $\uparrow$ & 97 & 73 & \underline{98} &  & 83 & 60 & \underline{90} &  & 42 & 30 & \underline{50} &  & \underline{98} & 72 & 90 &  & 77 & 56 & \underline{88} &  & 47 & 32 & \underline{52} \\ \hline
Ped. Col. $\downarrow$ & \textbf{0} & \textbf{0} & \textbf{0} &  & 4 & 7 & \textbf{2} &  & 24 & \textbf{16} & 18 &  & \textbf{0} & \textbf{0} & \textbf{0} &  & \textbf{2} & 4 & \textbf{2} &  & \textbf{12} & 16 & 20 \\
Veh. Col. $\downarrow$ & \textbf{0} & \textbf{0} & \textbf{0} &  & 8 & 9 & \textbf{2} &  & 18 & 33 & \textbf{11} &  & \textbf{0} & \textbf{0} & \textbf{0} &  & 17 & 18 & \textbf{3} &  & \textbf{26} & 34 & 30 \\
Other Col. $\downarrow$ & 2 & 10 & \textbf{1} &  & \textbf{5} & 12 & \textbf{5} &  & 13 & \textbf{5} & 15 &  & \textbf{0} & 8 & 6 &  & \textbf{3} & 12 & 6 &  & 11 & 7 & \textbf{6} \\
Time out $\downarrow$ & \textbf{1} & 17 & \textbf{1} &  & \textbf{0} & 12 & 1 &  & \textbf{3} & 16 & 6 &  & \textbf{2} & 20 & 4 &  & \textbf{1} & 10 & \textbf{1} &  & 4 & 11 & \textbf{2} \\ \bottomrule
\end{tabular}
}

\label{tabb:baseline}
\end{table*}

\begin{table}[t]
  \caption{\% of detected entities in features when the vehicle is stopped at green traffic light on NoCrash.}
  \centering
  \resizebox{0.6\linewidth}{!}{
  \begin{tabular}{@{}lc@{}}
    \toprule
    Cause of failure &  Percentage of detection\\
    \midrule
    \hline
    Red Traffic light & 56\% \\
    Pedestrian crossing & 18\% \\
    Vehicle obstruction & 3\%\\
    \textbf{Tot} & \textbf{77}\%\\
    \bottomrule
  \end{tabular}
  }
  \label{tab:faiss}

\end{table}

\begin{figure}[t]
\centering
\begin{tcolorbox}[colframe=red, left=0pt, right=-1pt,top=0pt, bottom=-2pt]
\includegraphics[width=.99\linewidth]{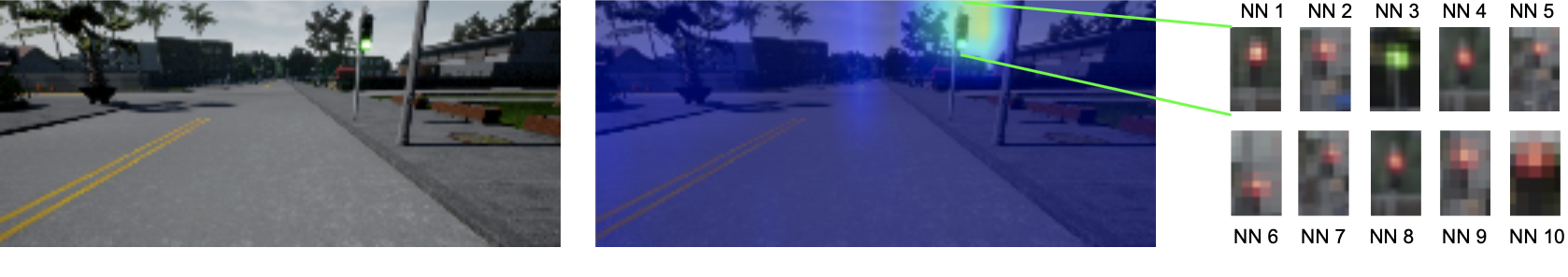} 
\includegraphics[width=.99\linewidth]{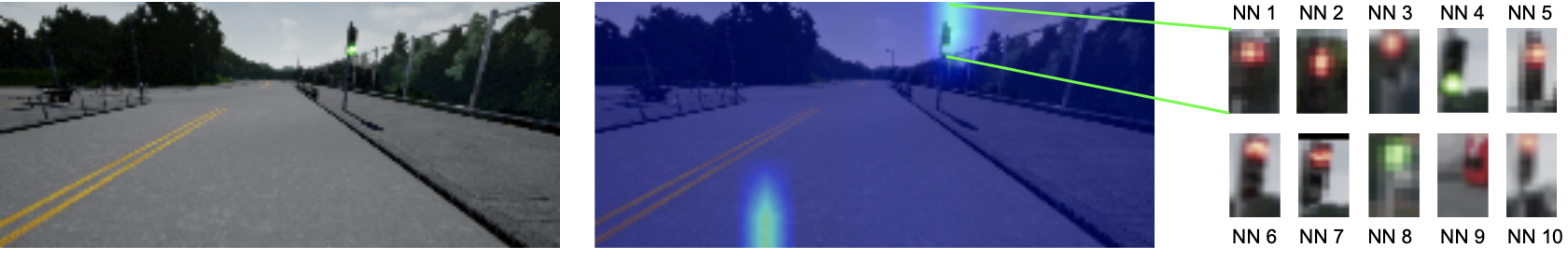} 
\end{tcolorbox}
\begin{tcolorbox}[colframe=green, left=0pt, right=-1pt,top=0pt, bottom=-2pt]
\includegraphics[width=.99\linewidth]{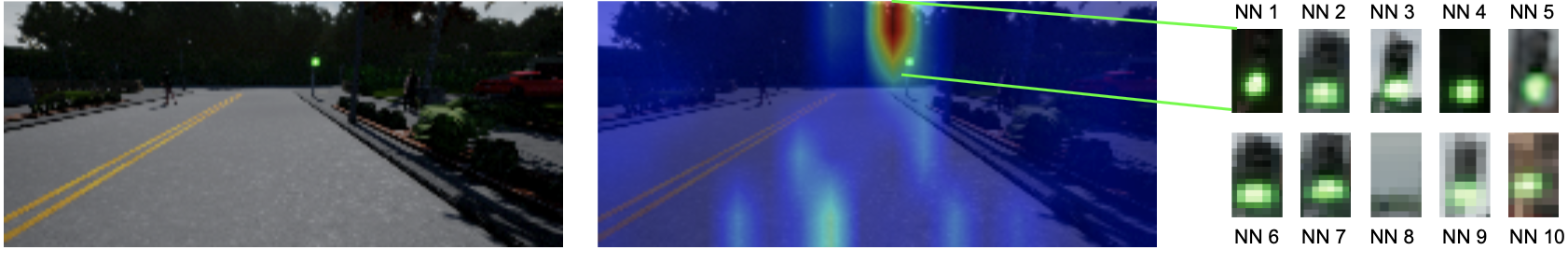} 
\end{tcolorbox}
\caption{Top 10 nighbors for the highest scoring attention after a traffic light turns green. We show examples of both successful crossing of the traffic light (framed in green) and failed due to red light "hallucination" (framed in red).}
\label{fig:heatmapDistribution}
\end{figure}

\section{State Token and Inertia Problem}
We inspect the relative importance of the state token and the image image content.
The state token emitted by the first stage is used to predict a stop/go decision with a dedicated loss. This makes the token carry useful information to the second stage, which is in charge of generating the actual driving commands.
In presence of a halt cue (e.g. red traffic light) \textcolor{black}{encoded in the state token propagated to the second stage}, the attention scheme of the second stage focuses on the state token rather than on the image patches.
When the state token indicates that the vehicle can advance, the attention focuses instead on the image patches to generate appropriate driving commands.
Fig. \ref{fig:StateToken} shows examples of stage 2 attention, with values of the state token and of the whole image, accumulated for each visual token.

The stop/go loss has a great impact on driving performance. In Tab. \ref{tab:ablations} we show the effect of removing such loss on the \textit{NoCrash} benchmark.
In densely trafficked environments, the success rate is almost halved when removing the loss.
Similar results are obtained with the single stage baseline.
We also test a model trained using a random vector as state token (w/o ST), yet keeping the stop/go loss: success rate heavily drops, especially in crowded environments. By feeding the state of the vehicle, the agent becomes aware of its speed and momentum, e.g. indicating whether and how a turn is taking place. This is hard to deduct from a single image.

Furthermore, the use of the state token and the stop/go loss, have a direct effect on addressing the inertia problem since the first stage is explicitly trained to predict movement.
Tab. \ref{tab:inertia} shows the failure rate due to inertia. As in \cite{codevilla2019exploring} we identify the inertia problem when an agent is still for 8 seconds before time out. Most failures of the single-stage baseline can be traced back to inertia and these are almost completely eliminated with the multi-stage model.
\textcolor{black}{Surprisingly, the NewWeather-Empty configuration in the NoCrash Benchmark exhibits the highest failure rate,  attributed to inertia (as indicated in Table 5). In this context, when the vehicle comes to a halt, it becomes trapped in a stationary state due to inertia. Notably, in an empty scenario, the sole discernible visual cue is the traffic light.
Conversely, in Regular or Dense scenarios, the dynamic nature of the environment allows the autonomous vehicle to break free from its stationary state by observing other vehicle behaviors and the dynamic surroundings, prompting a reevaluation of its decisions. In simple terms, the distance from a vehicle ahead or the dynamic behaviors of other agents can act as a trigger to escape from stall states caused by the inertia problem.}

A more general analysis of the causes of failure is also provided in Tab. \ref{tabb:baseline}. The multi-stage model considerably reduces collisions with pedestrians and vehicles, compared to the single-stage baseline. Interestingly failures due to time out (which include inertia) are almost eliminated.
Tab. \ref{tab:inertia} and Tab. \ref{tabb:baseline} indicate that, despite addressing in a very effective way the inertia problem, the model still suffers from a few inertia failures.
We exploit the Ex-post Semantic Explainability approach presented in Sec. \ref{sec:explainability} to inspect 50 episodes of the \textit{NoCrash} benchmark where the inertia problem still occurs at traffic lights.
In 56\% of the cases where the vehicle is stuck at a green light, the $k$ most similar features to the attended one contain a red traffic light, in 18\% a pedestrian crossing, and in 3\% a vehicle (Tab.~\ref{tab:faiss}).

In Fig. \ref{fig:heatmapDistribution} we show the top 10 nearest samples of the image region with the highest attention value \textcolor{black}{(first transformer stage)}. The first two rows show failure cases: the model correctly focuses on the traffic light but although it is green, the model maps it in a region of the latent space densely populated by red traffic lights.
We also show a sample of correct driving, where the vehicle accelerates as soon as the light turns green: retrieved images all depict green lights.
This suggests that what may appear as inertia might instead be confused with a failure of the backbone that mistakenly "hallucinates" halt cues.

\begin{figure}[t]
\centering
\includegraphics[width = \columnwidth,trim={0 0 10cm 0},clip]{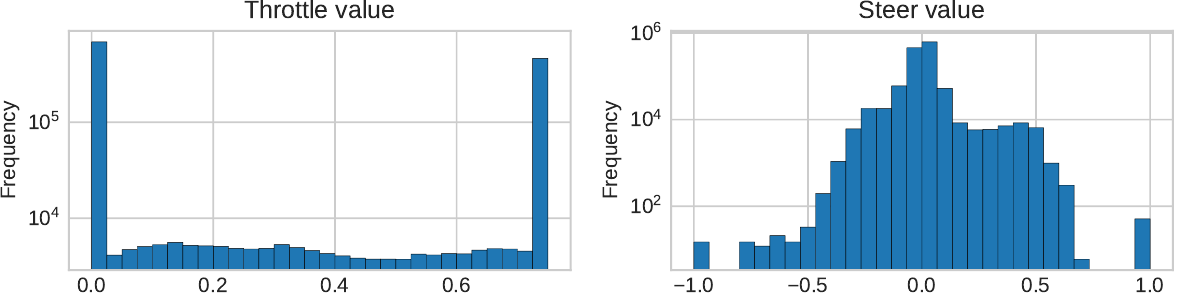}
  \caption{\textit{NoCrash} (left) Throttle distribution; (right) Steer distribution.}
  \label{fig:hists}
\end{figure}

\subsection{Online/Offline Evaluation and Noise Injection}
\label{sec:noiseinj}
\begin{figure*}[t]
 \centering
\includegraphics[width=0.75\columnwidth]{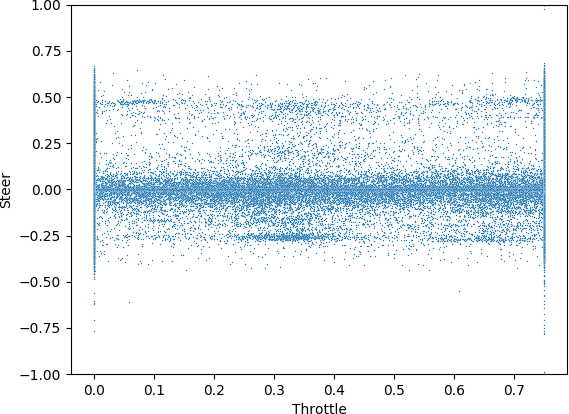} ~~~~~~~~~~~
\includegraphics[width=0.75\columnwidth]{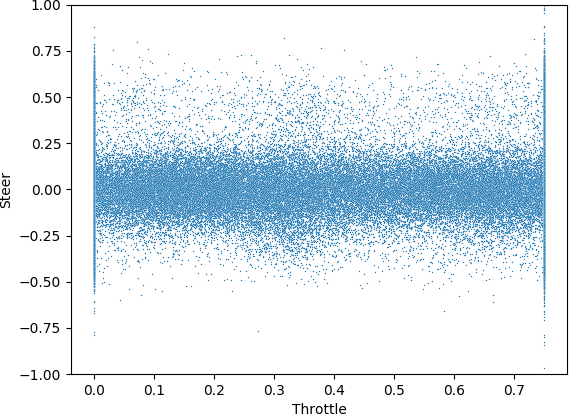} 
\caption{Distribution of Steer-Throttle without (left) and with (right) Noise Injection on \textit{NoCrash}. We have a better coverage of the space with Noise Injection.}
\label{fig:ThrottleDistribution}
\end{figure*}

To address the offline/online shift, exhaustive coverage at training time of possible input configurations (observed environment + internal state) could be a solution, yet it is difficult to achieve.
For instance, the \textit{NoCrash} dataset is unbalanced and throttle and steer values are extremely biased (Fig. \ref{fig:hists}). This limits the possibility of effectively inputting the vehicle state into the model at driving time.
Our data augmentation strategy that injects noise on the state token (Sec.~\ref{sec:state_token}) is intended to address this limitation.
We introduce a zero mean Gaussian noise with $\sigma=0.1$ on the driving controls (which are in $[0,1]$) and with $\sigma=1$ for speed. This has the effect of letting the model see at training time different combinations of state values.
In Fig. \ref{fig:ThrottleDistribution} we show the joint distribution of steer and throttle values with and without noise injection. Two modes for throttle can be observed corresponding to the over-represent stationary and full-throttle scenarios. At the same time, steer has a Gaussian distribution centered in zero (indicating no steer).
With noise injection, we get a more uniform distribution in the low-steer interval [-0.25, 0.25] across all throttle values. Also, higher steer values obtain a more uniform coverage.

\begin{figure}[t]
\centering
\includegraphics[width = 0.48\columnwidth]{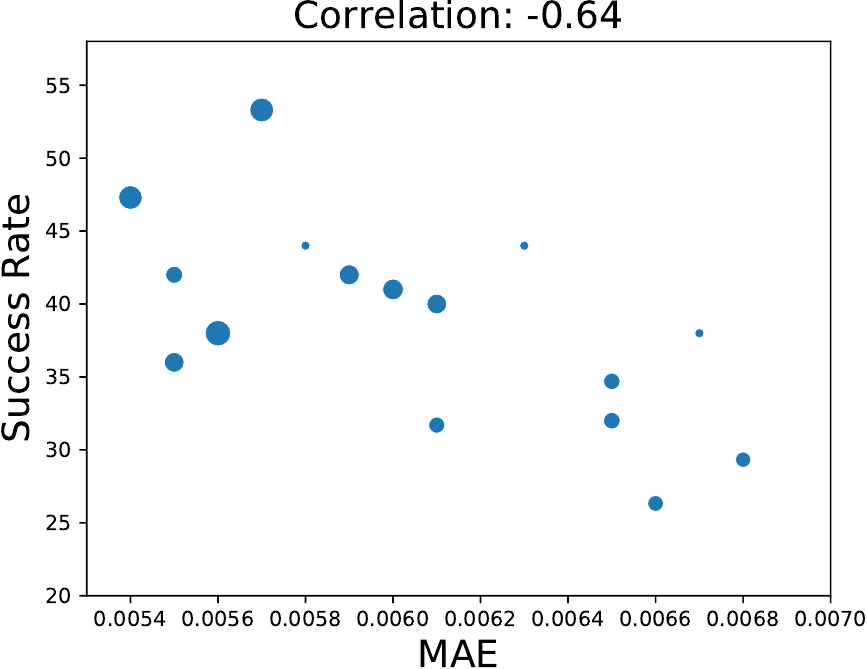}
\includegraphics[width = 0.48\columnwidth]{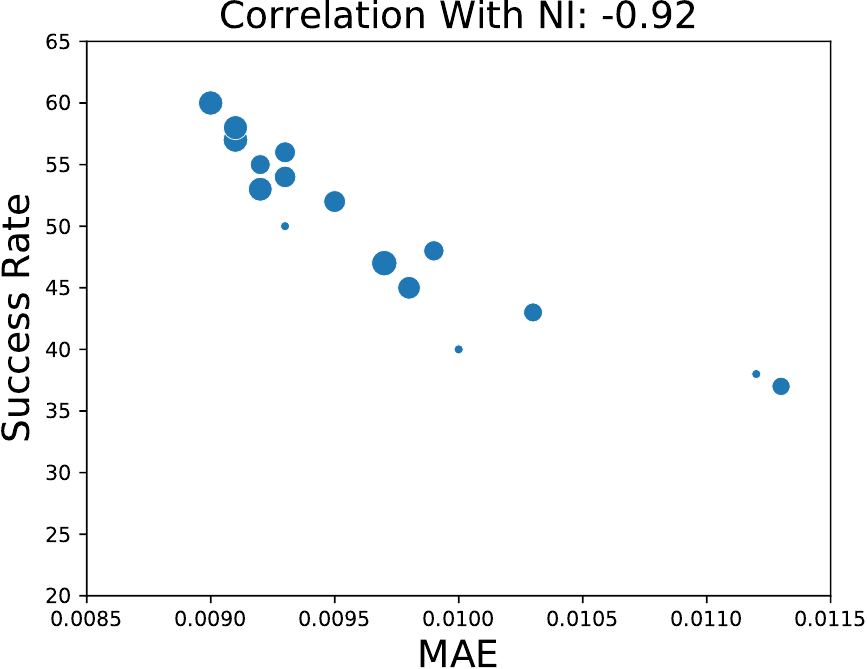}
  \caption{Pearson correlation between online success rate and offline validation MAE obtained by training the model multiple times without (left) and with (right) data augmentation on the state token. Dot size corresponds to different training epochs.}
  \label{fig:correlation}
\end{figure}

In Fig. \ref{fig:correlation} we quantify the correlation between online success rate and offline validation MAE using the sample Pearson correlation coefficient, as done in \cite{codevilla2018offline}. We plot the results without using data augmentation via noise injection (corr: -0.64) and with (corr: -0.92). Despite not having a huge impact on the results in training conditions, as shown in Tab. \ref{tab:ablations}, in generalization conditions noise injection brings noticeable benefits.
From the plots in Fig. \ref{fig:correlation} it can be seen that without using data augmentation there are huge differences for similar MAE values (e.g., 20\% success rate gap with a small difference of 0.0001 in MAE).

Another component to help the agent act as the expert demonstrator is the command consistency module (see \ref{sec:CCM}). This module acts as a regularizer, encouraging the model to generate driving commands that are not in conflict with each other, and thus preventing unwanted behaviors at driving time.
The necessity of CCM also stems from the fact that maneuvers (e.g. a right turn) could be performed in different ways (e.g. slow and narrow or fast and wide turn). Results in Tab.~\ref{tab:ablations} confirm the usefulness of CCM.


\subsection{On the Command Coherency Module}
\begin{figure}[t]
 \centering
\includegraphics[width=\columnwidth]{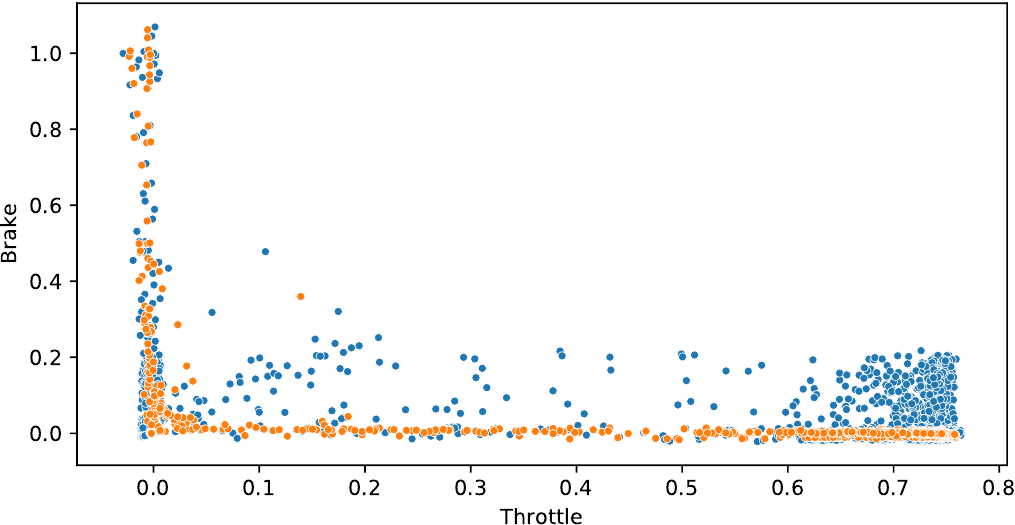}
\caption{Distribution of Brake-Throttle values with the Command Coherency Module disabled (blue) and enabled (orange).}
\label{fig:Brake_vs_Throttle_CCM}
\end{figure}

In the proposed architecture, the CCM module is responsible for the generation of non-conflicting throttle and brake commands at training time. 
The impact of this module on the coherency of throttle and brake pairs and ultimately on the accuracy of driving at test time is demonstrated in the experiments reported in Table \ref{tab:ablations} that highlight a severe performance drop when training is conducted with the CCM disabled.
To delve deeper into the effect of the CCM module, we log pairs of throttle and brake command values during a driving session executed twice, with CCM respectively enabled and disabled.
For this experiment, we use an episode of the \textit{NoCrash} benchmark consisting of about 3000 frames. 
Values of pairs (throttle, brake) are shown in the scatter plots of Fig.~\ref{fig:Brake_vs_Throttle_CCM}. 
When the CCM is disabled, it can be noticed that a relevant portion of the outputs is characterized by both throttle and brake values greater than zero, meaning that the vehicle is both trying to accelerate and decelerate at the same time.
By using the CCM, almost all these non-coherent configurations disappear and only one of the two commands at a time can take values significantly greater than zero.

\section{Conclusions and Future Works}
\label{sec:conclusions}
We addressed two major issues in training a state-aware model with imitation learning. First, the inertia problem has been dealt with using a multi-stage architecture with state token propagation, where the first stage learns to inform the next one about stop/go decisions. We report extremely low rates of inertia. Second, the offline/online gap has been bridged by performing data augmentation on the state token, significantly increasing the correlation between success rate and validation error. In addition, we also exploited built-in visual attention with a retrieval-based ex-post explainability to characterize failures. We found that what may appear as inertia might indeed be caused by a completely different problem, such as backbone hallucinations.
In future works, we intend to exploit the hierarchical structure of the model to create an inspection chain to debug the model: whenever the explainability allows us to discover an issue, we can add a new level in the hierarchy to enrich the state token with new information and condition the generation of the driving commands. An interesting approach in this direction would be to study a model with different modules to be deployed in parallel rather than stacking them hierarchically.
Furthermore, the proposed model could be easily improved to include additional sources of data such as depth and segmentation, e.g. concatenating them to the input image.

{\small
\bibliographystyle{IEEEtran}
\bibliography{egbib}

\begin{thebibliography}{10}
\providecommand{\url}[1]{#1}
\csname url@samestyle\endcsname
\providecommand{\newblock}{\relax}
\providecommand{\bibinfo}[2]{#2}
\providecommand{\BIBentrySTDinterwordspacing}{\spaceskip=0pt\relax}
\providecommand{\BIBentryALTinterwordstretchfactor}{4}
\providecommand{\BIBentryALTinterwordspacing}{\spaceskip=\fontdimen2\font plus
\BIBentryALTinterwordstretchfactor\fontdimen3\font minus
  \fontdimen4\font\relax}
\providecommand{\BIBforeignlanguage}[2]{{%
\expandafter\ifx\csname l@#1\endcsname\relax
\typeout{** WARNING: IEEEtran.bst: No hyphenation pattern has been}%
\typeout{** loaded for the language `#1'. Using the pattern for}%
\typeout{** the default language instead.}%
\else
\language=\csname l@#1\endcsname
\fi
#2}}
\providecommand{\BIBdecl}{\relax}
\BIBdecl

\bibitem{pohlen2017full}
T.~Pohlen, A.~Hermans, M.~Mathias, and B.~Leibe, ``Full-resolution residual
  networks for semantic segmentation in street scenes,'' in \emph{Proceedings
  of the IEEE conference on computer vision and pattern recognition}, 2017, pp.
  4151--4160.

\bibitem{claussmann2019review}
L.~Claussmann, M.~Revilloud, D.~Gruyer, and S.~Glaser, ``A review of motion
  planning for highway autonomous driving,'' \emph{IEEE Transactions on
  Intelligent Transportation Systems}, vol.~21, no.~5, pp. 1826--1848, 2019.

\bibitem{marchetti2020multiple}
F.~Marchetti, F.~Becattini, L.~Seidenari, and A.~Del~Bimbo, ``Multiple
  trajectory prediction of moving agents with memory augmented networks,''
  \emph{IEEE Transactions on Pattern Analysis and Machine Intelligence}, 2020.

\bibitem{codevilla2018end}
F.~Codevilla, M.~Müller, A.~López, V.~Koltun, and A.~Dosovitskiy,
  ``End-to-end driving via conditional imitation learning,'' \emph{In 2018 IEEE
  International Conference on Robotics and Automation (ICRA)}, pp. 1--9, 2018.

\bibitem{codevilla2019exploring}
F.~Codevilla, E.~Santana, A.~M. L{\'o}pez, and A.~Gaidon, ``Exploring the
  limitations of behavior cloning for autonomous driving,'' in \emph{Proc. of
  the IEEE/CVF International Conference on Computer Vision}, 2019, pp.
  9329--9338.

\bibitem{de2019causal}
P.~De~Haan, D.~Jayaraman, and S.~Levine, ``Causal confusion in imitation
  learning,'' \emph{Advances in Neural Information Processing Systems},
  vol.~32, 2019.

\bibitem{greco2022imitation}
A.~Greco, L.~Rundo, A.~Saggese, M.~Vento, and A.~Vicinanza, ``Imitation
  learning for autonomous vehicle driving: How does the representation
  matter?'' in \emph{International Conference on Image Analysis and
  Processing}.\hskip 1em plus 0.5em minus 0.4em\relax Springer, 2022, pp.
  15--26.

\bibitem{samsami2021causal}
M.~R. Samsami, M.~Bahari, S.~Salehkaleybar, and A.~Alahi, ``Causal imitative
  model for autonomous driving,'' \emph{arXiv preprint arXiv:2112.03908}, 2021.

\bibitem{le2022survey}
L.~Le~Mero, D.~Yi, M.~Dianati, and A.~Mouzakitis, ``A survey on imitation
  learning techniques for end-to-end autonomous vehicles,'' \emph{IEEE
  Transactions on Intelligent Transportation Systems}, 2022.

\bibitem{codevilla2018offline}
F.~Codevilla, A.~M. Lopez, V.~Koltun, and A.~Dosovitskiy, ``On offline
  evaluation of vision-based driving models,'' in \emph{Proceedings of the
  European Conference on Computer Vision (ECCV)}, 2018, pp. 236--251.

\bibitem{ross2010efficient}
S.~Ross and D.~Bagnell, ``Efficient reductions for imitation learning,'' in
  \emph{Proceedings of the thirteenth international conference on artificial
  intelligence and statistics}.\hskip 1em plus 0.5em minus 0.4em\relax JMLR
  Workshop and Conference Proceedings, 2010, pp. 661--668.

\bibitem{ross2011reduction}
S.~Ross, G.~Gordon, and D.~Bagnell, ``A reduction of imitation learning and
  structured prediction to no-regret online learning,'' in \emph{Proceedings of
  the fourteenth international conference on artificial intelligence and
  statistics}, 2011, pp. 627--635.

\bibitem{schroecker2017state}
Y.~Schroecker and C.~L. Isbell, ``State aware imitation learning,''
  \emph{Advances in Neural Information Processing Systems}, 2017.

\bibitem{dosovitskiy2020image}
A.~Dosovitskiy, L.~Beyer, A.~Kolesnikov, D.~Weissenborn, X.~Zhai,
  T.~Unterthiner, M.~Dehghani, M.~Minderer, G.~Heigold, S.~Gelly \emph{et~al.},
  ``An image is worth 16x16 words: Transformers for image recognition at
  scale,'' \emph{arXiv preprint arXiv:2010.11929}, 2020.

\bibitem{vaswani2017attention}
A.~Vaswani, N.~Shazeer, N.~Parmar, J.~Uszkoreit, L.~Jones, A.~N. Gomez,
  {\L}.~Kaiser, and I.~Polosukhin, ``Attention is all you need,''
  \emph{Advances in neural information processing systems}, vol.~30, 2017.

\bibitem{omeiza2021explanations}
D.~Omeiza, H.~Webb, M.~Jirotka, and L.~Kunze, ``Explanations in autonomous
  driving: A survey,'' \emph{IEEE Trans. on Intelligent Transportation
  Systems}, 2021.

\bibitem{cultrera2020explaining}
L.~Cultrera, L.~Seidenari, F.~Becattini, P.~Pala, and A.~Del~Bimbo,
  ``Explaining autonomous driving by learning end-to-end visual attention,'' in
  \emph{Proceedings of the IEEE/CVF Conference on Computer Vision and Pattern
  Recognition Workshops}, 2020, pp. 340--341.

\bibitem{zablocki2021explainability}
{\'E}.~Zablocki, H.~Ben-Younes, P.~P{\'e}rez, and M.~Cord, ``Explainability of
  vision-based autonomous driving systems: Review and challenges,'' \emph{arXiv
  preprint arXiv:2101.05307}, 2021.

\bibitem{argall2009survey}
B.~D. Argall, S.~Chernova, M.~Veloso, and B.~Browning, ``A survey of robot
  learning from demonstration,'' \emph{Robotics and autonomous systems},
  vol.~57, no.~5, pp. 469--483, 2009.

\bibitem{attia2018global}
A.~Attia and S.Dayan, ``Global overview of imitation learning,'' \emph{arXiv
  1801.06503v1}, 2018.

\bibitem{bojarski2016end}
M.~Bojarski, D.~Del~Testa, D.~Dworakowski, B.~Firner, B.~Flepp, P.~Goyal, L.~D.
  Jackel, M.~Monfort, U.~Muller, J.~Zhang \emph{et~al.}, ``End to end learning
  for self-driving cars,'' \emph{arXiv preprint arXiv:1604.07316}, 2016.

\bibitem{sauer2018conditional}
A.~Sauer, N.~Savi-nov, and A.~Geiger, ``Conditional affordance learning for
  driving in urban environments,'' \emph{in Conference on Robot Learning
  (CoRL)}, 2018.

\bibitem{xiao2020multimodal}
Y.~Xiao, F.~Codevilla, A.~Gurram, O.~Urfalioglu, and A.~M. L{\'o}pez,
  ``Multimodal end-to-end autonomous driving,'' \emph{IEEE Transactions on
  Intelligent Transportation Systems}, vol.~23, no.~1, pp. 537--547, 2020.

\bibitem{haris2022navigating}
M.~Haris and A.~Glowacz, ``Navigating an automated driving vehicle via the
  early fusion of multi-modality,'' \emph{Sensors}, vol.~22, no.~4, p. 1425,
  2022.

\bibitem{liang2018cirl}
X.~Liang, T.~Wang, and E.~X. L.~Yang, ``Cirl: Controllable imitative
  reinforcement learning for vision-based self-driving,'' \emph{in European
  Conference on Computer Vision (ECCV)}, 2018.

\bibitem{toromanoff2020end}
M.~Toromanoff, E.~Wirbel, and F.~Moutarde, ``End-to-end model-free
  reinforcement learning for urban driving using implicit affordances,'' in
  \emph{Proceedings of the IEEE/CVF conference on computer vision and pattern
  recognition}, 2020, pp. 7153--7162.

\bibitem{zhang2021end}
Z.~Zhang, A.~Liniger, D.~Dai, F.~Yu, and L.~Van~Gool, ``End-to-end urban
  driving by imitating a reinforcement learning coach,'' in \emph{Proceedings
  of the IEEE/CVF international conference on computer vision}, 2021, pp.
  15\,222--15\,232.

\bibitem{lee2018context}
D.~Lee, S.~Liu, J.~Gu, M.-Y. Liu, M.-H. Yang, and J.~Kautz, ``Context-aware
  synthesis and placement of object instances,'' \emph{Advances in neural
  information processing systems}, vol.~31, 2018.

\bibitem{berlincioni2021multiple}
L.~Berlincioni, F.~Becattini, L.~Seidenari, and A.~Del~Bimbo, ``Multiple future
  prediction leveraging synthetic trajectories,'' in \emph{2020 25th
  International Conference on Pattern Recognition (ICPR)}.\hskip 1em plus 0.5em
  minus 0.4em\relax IEEE, 2021, pp. 6081--6088.

\bibitem{chen2020learning}
D.~Chen, B.~Zhou, V.~Koltun, and P.~Kr{\"a}henb{\"u}hl, ``Learning by
  cheating,'' in \emph{Conference on Robot Learning}.\hskip 1em plus 0.5em
  minus 0.4em\relax PMLR, 2020, pp. 66--75.

\bibitem{yang2018end}
Z.~Yang, Y.~Zhang, J.~Yu, J.~Cai, and J.~Luo, ``End-to-end multi-modal
  multi-task vehicle control for self-driving cars with visual perceptions,''
  in \emph{2018 24th International Conference on Pattern Recognition
  (ICPR)}.\hskip 1em plus 0.5em minus 0.4em\relax IEEE, 2018, pp. 2289--2294.

\bibitem{li2018ret}
Z.~Li, T.~Motoyoshi, T.~O. K.~Sasaki, and S.~Sugano, ``Rethinking self-driving:
  Multi-task knowledge for better generalization and accident explanation
  ability,'' \emph{In Proc. of the European Conference on Computer Vision
  (ECCV)}, 2018.

\bibitem{chekroun2021gri}
R.~Chekroun, M.~Toromanoff, S.~Hornauer, and F.~Moutarde, ``Gri: General
  reinforced imitation and its application to vision-based autonomous
  driving,'' \emph{arXiv preprint arXiv:2111.08575}, 2021.

\bibitem{chen2022learning}
D.~Chen and P.~Kr{\"a}henb{\"u}hl, ``Learning from all vehicles,'' in
  \emph{Proceedings of the IEEE/CVF Conference on Computer Vision and Pattern
  Recognition}, 2022, pp. 17\,222--17\,231.

\bibitem{chen2021learning}
D.~Chen, V.~Koltun, and P.~Kr{\"a}henb{\"u}hl, ``Learning to drive from a world
  on rails,'' in \emph{Proceedings of the IEEE/CVF International Conference on
  Computer Vision}, 2021, pp. 15\,590--15\,599.

\bibitem{chitta2021neat}
K.~Chitta, A.~Prakash, and A.~Geiger, ``Neat: Neural attention fields for
  end-to-end autonomous driving,'' in \emph{Proceedings of the IEEE/CVF
  International Conference on Computer Vision}, 2021, pp. 15\,793--15\,803.

\bibitem{hu2022st}
S.~Hu, L.~Chen, P.~Wu, H.~Li, J.~Yan, and D.~Tao, ``St-p3: End-to-end
  vision-based autonomous driving via spatial-temporal feature learning,'' in
  \emph{Computer Vision--ECCV 2022: 17th European Conference, Tel Aviv, Israel,
  October 23--27, 2022, Proceedings, Part XXXVIII}.\hskip 1em plus 0.5em minus
  0.4em\relax Springer, 2022, pp. 533--549.

\bibitem{chitta2022transfuser}
K.~Chitta, A.~Prakash, B.~Jaeger, Z.~Yu, K.~Renz, and A.~Geiger, ``Transfuser:
  Imitation with transformer-based sensor fusion for autonomous driving,''
  \emph{arXiv preprint arXiv:2205.15997}, 2022.

\bibitem{chen2015deep}
C.~Chen, A.~Steff, A.~Kornhauser, and J.~Xiao, ``Deepdriving: Learning
  affordance for direct perception in autonomous drivings,'' \emph{In Proc. of
  the IEEE International Conference on Computer Vision}, pp. 2722--2730, 2015.

\bibitem{cultrera2023explaining}
L.~Cultrera, F.~Becattini, L.~Seidenari, P.~Pala, and A.~Del~Bimbo,
  ``Explaining autonomous driving with visual attention and end-to-end
  trainable region proposals,'' \emph{Journal of Ambient Intelligence and
  Humanized Computing}, pp. 1--13, 2023.

\bibitem{xiao2023scaling}
Y.~Xiao, F.~Codevilla, D.~P. Bustamante, and A.~M. Lopez, ``Scaling
  self-supervised end-to-end driving with multi-view attention learning,''
  \emph{arXiv preprint arXiv:2302.03198}, 2023.

\bibitem{pomerleau1988alvinn}
D.~A. Pomerleau, ``Alvinn: An autonomous land vehicle in a neural network,''
  \emph{Advances in neural information processing systems}, vol.~1, 1988.

\bibitem{chuang2022resolving}
C.-C. Chuang, D.~Yang, C.~Wen, and Y.~Gao, ``Resolving copycat problems in
  visual imitation learning via residual action prediction,'' in \emph{Computer
  Vision--ECCV 2022: 17th European Conference, Tel Aviv, Israel, October
  23--27, 2022, Proceedings, Part XXXIX}.\hskip 1em plus 0.5em minus
  0.4em\relax Springer, 2022, pp. 392--409.

\bibitem{johnson2019billion}
J.~Johnson, M.~Douze, and H.~J{\'e}gou, ``Billion-scale similarity search with
  gpus,'' \emph{IEEE Transactions on Big Data}, vol.~7, no.~3, pp. 535--547,
  2019.

\bibitem{dosovitskiy2017carla}
A.~Dosovitskiy, G.~Ros, F.~Codevilla, and A.~López, ``Carla: An open urban
  driving simulator,'' \emph{In Conference on Robot Learning (CoRL)}, 2017.

\bibitem{ishihara2021multi}
K.~Ishihara, A.~Kanervisto, J.~Miura, and V.~Hautamaki, ``Multi-task learning
  with attention for end-to-end autonomous driving,'' in \emph{Proc. of the
  IEEE/CVF Conference on Computer Vision and Pattern Recognition}, 2021, pp.
  2902--2911.

\bibitem{kim2020multi}
I.~Kim, H.~Lee, J.~Lee, E.~Lee, and D.~Kim, ``Multi-task learning with future
  states for vision-based autonomous driving,'' in \emph{Proceedings of the
  Asian Conference on Computer Vision}, 2020.

\bibitem{kim2022learning}
I.~Kim, J.~Lee, and D.~Kim, ``Learning mixture of domain-specific experts via
  disentangled factors for autonomous driving,'' in \emph{Proceedings of the
  AAAI Conference on Artificial Intelligence}, vol.~36, no.~1, 2022, pp.
  1148--1156.

\bibitem{zhao2022cadre}
Y.~Zhao, K.~Wu, Z.~Xu, Z.~Che, Q.~Lu, J.~Tang, and C.~H. Liu, ``Cadre: A
  cascade deep reinforcement learning framework for vision-based autonomous
  urban driving,'' in \emph{Proceedings of the AAAI Conference on Artificial
  Intelligence}, vol.~36, no.~3, 2022, pp. 3481--3489.

\end{thebibliography}
}

\vspace{-33pt}
\begin{IEEEbiographynophoto}{Luca Cultrera} received a PhD degree in information engineering  from the University of Florence in 2023.
Currently he is post-doctoral fellow at Media Integration and Communication Center (MICC) at the University of Florence.
\end{IEEEbiographynophoto}

\vspace{-33pt}
\begin{IEEEbiographynophoto}{Federico Becattini} is a Tenure-Track Assistant Professor at the University of Siena. His research focuses on computer vision and memory-based learning. He organized tutorials and workshops at ICPR2020, ICIAP2020, ACMMM2022, ICPR2022, ECCV2022. He has co-authored more than 40 papers. He is Associate Editor of the International Journal of Multimedia Information Retrieval (IJMIR).
\end{IEEEbiographynophoto}

\vspace{-33pt}
\begin{IEEEbiographynophoto}{Lorenzo Seidenari} is an Associate Professor at the University of Florence. His research focuses on deep learning for object and action recognition. He is an ELLIS scholar. He was a visiting scholar at the University of Michigan in 2013. He gave a tutorial at ICPR 2012 on image categorization. He is author of 16 journal papers and more than 40 peer-reviewed conference papers.
\end{IEEEbiographynophoto}
\vspace{-33pt}

\begin{IEEEbiographynophoto}{Pietro Pala} is full professor of computer science and engineering at the University of Florence.
His research activity focuses on 3D data processing with deep learning models for biometrics and action recognition. He is author of over 40 journal papers and over 140 peer-reviewed conference papers. He is associate editor of the ACM Transactions on Multimedia Computing, Communications, and Applications, and of the Springer Multimedia Systems.
\end{IEEEbiographynophoto}

\vspace{-33pt}
\begin{IEEEbiographynophoto}{Alberto Del Bimbo} is an Emeritus Professor and was director of the Media Integration and Communication Center at University of Florence. His interests are multimedia vision. He was Co-Chair of ACMMM2010, ECCV2012, ICPR2020. ACM nominated him Distinguished Scientist and he received the SIGMM Technical Achievement Award for Outstanding Technical Contributions to Multimedia Computing, Communications and Applications.
\end{IEEEbiographynophoto}

\vfill

\end{document}